\documentclass{bmvc2k}

\usepackage{cleveref}
\usepackage{booktabs}
\usepackage{xcolor, colortbl}
\usepackage[rightcaption]{sidecap}
\usepackage{amssymb} 

\usepackage{capt-of}
\usepackage{multirow}
\usepackage{subfigure}
\usepackage[parfill]{parskip}
\newcommand{\midsepremove}{\aboverulesep = 0.1mm \belowrulesep = 0.2mm}
\midsepremove
\newcommand{\midsepdefault}{\aboverulesep = 0.605mm \belowrulesep = 0.984mm}
\midsepdefault
\midsepremove

\title{Prompt Generation Networks for Input-Space Adaptation of Frozen Vision Transformers}

\addauthor{Jochem Loedeman}{j.m.loedeman@gmail.com}{1}
\addauthor{Maarten C. Stol}{maarten.stol@braincreators.com}{2}
\addauthor{Tengda Han}{htd@robots.ox.ac.uk}{3}
\addauthor{Yuki M. Asano}{y.m.asano@uva.nl}{1}

\addinstitution{
University of Amsterdam
}
\addinstitution{
BrainCreators
}
\addinstitution{
University of Oxford
}

\runninghead{Loedeman et al.}{Prompt Generation Networks}

\def\eg{\emph{e.g}\bmvaOneDot}

\def\etal{\emph{et al}\bmvaOneDot}

\begin{document}

\maketitle

\begin{abstract}
With the introduction of the transformer architecture in computer vision, increasing model scale has been demonstrated as a clear path to achieving performance and robustness gains. However, with model parameter counts reaching the billions, classical finetuning approaches are becoming increasingly limiting and even unfeasible when models become hosted as inference APIs, as in NLP. Visual input-prompt learning, an adaptation technique in which additional inputs in visual (RGB) space are learned, has emerged as a potential solution for adapting frozen and cloud-hosted models, requiring neither access to the forward pass, nor post-processing.
Yet so far, these constraints have deteriorated adaptation performances significantly. To this end, we propose the Prompt Generation Network (PGN) that generates a different prompt for every data point, which is then used to adapt a frozen pretrained vision model to a target task.
We show that the PGN effectively adapts pretrained models to various new datasets: It surpasses previous methods by a large margin on 12/12 datasets and even outperforms full-finetuning on 5/12, while requiring 100x fewer parameters.
Lastly, we introduce the ``prompt inversion'' trick, with which PGNs can be efficiently trained in a latent space but deployed in RGB input space for inference. 
\end{abstract}

%-------------------------------------------------------------------------
\section{Introduction}
Large-scale pretrained models, such as those obtained from self-supervised or vision-language pretraining have shown remarkable performance gains for various visual tasks.
Particularly models such as SEER~\cite{goyal2021self}, CLIP~\citep{Radford2021} and ALIGN~\citep{Jia2021} have demonstrated that large and multi-modal datasets can yield models that exhibit novel abilities in terms of robustness or few- and zero-shot learning. 
However, the requirements in terms of dataset size and compute infrastructure are exceedingly prohibitive for most researchers, such that these models are and will likely remain limited in terms of their diversity and availability. 
Exacerbating this trend further, model sizes are steadily increasing: a case in point is the recent development of a Vision Transformer model with 22B parameters~\cite{dehghani2023scaling}. 
Eventually, models will reach sizes that can only be served via dedicated hardware or API interfaces, as is already the case for the largest and best models in NLP (\eg~Llama 3~\citep{dubey2024llama}).

Both of these developments mean that many classic approaches for finetuning models are not applicable anymore.
This is because the prevailing paradigms entangle the adaptation and computation phase, for example by directly adapting weights in a model or by requiring internal access to the model's forward pass and modifying it with additional inputs~\cite{Jia2021}, or additional intermediate layers~\cite{zhang2021tip}.
A solution to this is adapting frozen models by solely learning additional inputs.
This approach covers not only the setting where the model is impossible to directly change, \textit{e.g.} because it is served via an API or hard-coded as an application-specific integrated circuit (ASIC), but also where it is very sensitive to changes, \textit{e.g.} due to quantization.
% called \textit{prompts}.
Additionally learned inputs, also called \textit{prompts}, are typically learned per domain and downstream task and have shown promising results for adapting to image domains~\citep{Bahng2022} and are broadly inspired from adversarial reprogramming~\cite{elsayed2018adversarial, kloberdanz2021improved}.
However, so far, the resulting performances have fallen short compared to finetuning models.

In this paper, we argue that the main reason for this shortfall is the arbitrary definition of a domain and the resulting limited modeling flexibility, as the prompt set is constant and therefore does not depend on the input data.
In contrast, we propose a new method that allows adapting to \emph{every single} input image. 
For this, we introduce the Prompt Generation Network (PGN) that learns to generate new prompts for every image by combining items from a jointly learned library of tokens.
Furthermore, as these prompts' purpose is to only aid the large-scale model, the PGN can be kept lightweight, which allows for efficient, yet strong modelling capabilities. 
By fully decoupling the adaptation part from the internal model computations, a PGN can be simply deployed on client devices after an initial stage of server-side training.
On a benchmark covering 12 datasets, we show that our PGN approach achieves performances that matches and outperforms those of fully finetuned models, while being two orders of magnitude more efficient in terms of additional parameters. 
This demonstrates that our proposed method effectively closes the divide between high-performing classical finetuning and previously constrained adaptation of frozen vision models.

Overall, this paper makes four main contributions:
\begin{itemize}
  \setlength\itemsep{0.01em}
    \item We develop a simple and effective framework for learning input-dependent visual prompts via Prompt Generation Networks.
    \item We propose an alternative inference mode that decouples the PGN from the large-scale model, such that it is possible to arrange them in a client-server setup, making our approach compatible with recent developments in the industry.
    \item We demonstrate the generalizability and state-of-the-art performances of the proposed method across 12 datasets, architectures and settings such as multi-dataset inference.
    \item Finally, via quantitative and qualitative analyses, we showcase how with our method a ``division of labor'' emerges  between the frozen model and the PGN.
\end{itemize}

\section{Related works}

While pretrained vision models generally need to be finetuned, vision-language models can perform zero-shot transfer by prompting the text encoder to perform downstream tasks.
However, for most datasets there is still a significant performance gap with respect to full-finetuning, which is computationally expensive and reduces robustness~\cite{wortsman2022robust}. 
A range of alternative methods has been proposed to address these issues. 

\paragraph{Adapters and partial finetuning.}
Inspired by previous works in NLP~\citep{houlsby2019parameter, pfeiffer2020adapterhub}, lightweight feature adapters (made of trainable parts in between frozen ones) have shown performance gains~\citep{gao2021clip, zhang2021tip}. 
Requiring few additional parameters, only finetuning the bias parameters of pretrained models has also been proposed~\cite{zaken2021bitfit,Han2020tiny}.
In contrast, VPT~\citep{Jia2022} learns additional inputs to multiple layers in a pretrained vision transformer~\cite{Dosovitskiy2020} but also finetunes a linear classifier on top. 
Another finetuning-based approach proposed in~\citep{wortsman2022robust} is to ensemble the weights between zero-shot and finetuned models and which trains additional networks that are fused via summation~\citep{zhang2020side}.
Specific to vision-language models,~\citep{Zhou2022} learn an adaptation network between CLIP's vision and text encoders.

Although these works show promising results, they either change the model's weights or require access to the internals of the forward pass for inserting additional computations. % of the modelfor the insertion of prompt vectors.
This is a key difference to our work, in which we strictly separate the adaptation stage from the frozen model computation stage.
\begin{figure*}[t!]
\begin{center}
\includegraphics[width=\textwidth]{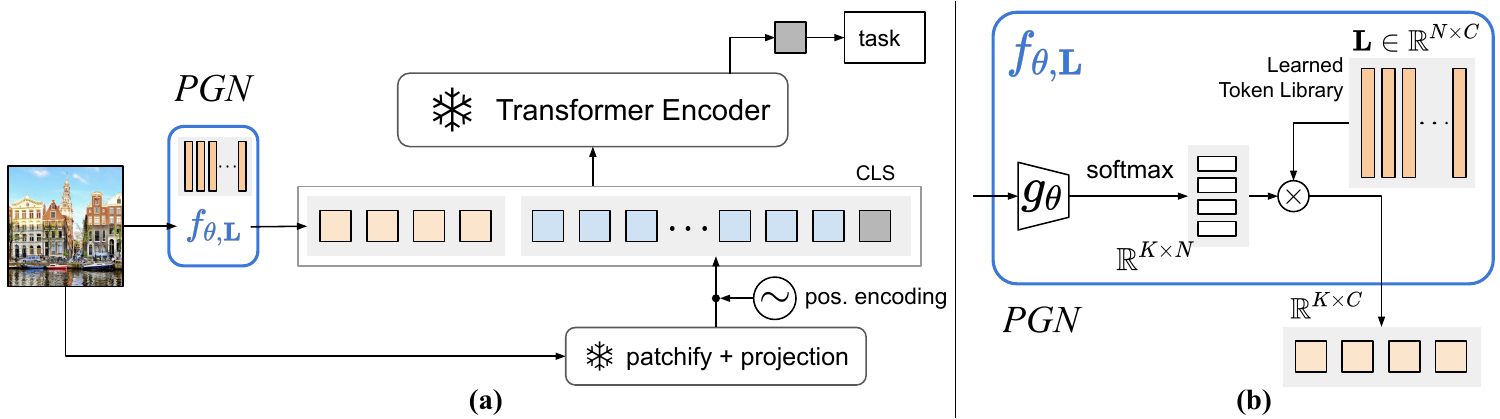}
\end{center}
\caption{
We propose the~\textbf{Prompt Generation Network (PGN)}, 
a simple yet effective method that generates prompts conditioned 
on the input images that benefits the domain adaptation process, while keeping the whole pretrained transformer frozen. 
\textbf{(a)}: the overall learning pipeline including PGN, denoted by $f_{\theta,\mathbf{L}}$:
the learned prompt vectors are fed into the pretrained transformer encoder together with image patches
for the task or domain of interest.
\textbf{(b)}: the detailed structure of PGN, a lightweight neural network $g_\theta$ learns probability distributions to select prompt vectors from a Token Library~$\mathbf{L}$.
}
\label{fig:method}
\end{figure*}
\paragraph{Input-space adaptation.} % prompt learning and reprogramming.
Optimizing solely in the \textit{input-space}, prompt learning originates from language modeling and is a lightweight approach to adapting a model to perform downstream tasks~\citep{Radford2019, Brown2020, LiLiang2021}. 
For computer vision, initial efforts focused on learning continuous prompts for the text encoder of pretrained vision-language models for image~\citep{Zhou2021,yao2021cpt} and video tasks~\citep{Ju2021}. 
In contrast to these, our method does not rely on multi-modal systems or a general text encoder but can be used for any visual encoder.

\noindent Most related to our work is the work of~\citep{Bahng2022}, that proposes visual prompting through the addition of learnable pixels, effectively learning prompts in the data space. 
This is the same setting as adversarial reprogramming~\citep{elsayed2018adversarial, kloberdanz2021improved}, where pretrained CNNs are repurposed to classify images from different datasets by applying transformations \textit{only to the input}. This form of adaptation therefore does not affect the model itself, making it an excellent option for scenarios with restricted model access during inference.

And although these works show significant gains, they do not match the performance of classic adaptation techniques like finetuning and linear probing.
Our method bridges the performance gap by learning prompts that, unlike previous works, are adapted to \mbox{each image}.
\section{Methods}
\subsection{Review of prompt learning methods}
\label{method:vp}

In NLP, prompt learning offers a lightweight approach for tuning pretrained models for performing downstream tasks.
Let $\Phi_{T}(.)$ denote the pretrained language model
and $\mathbf{x}_T = [ a_1, a_2, ..., a_n ]$ denote the
input language tokens.
Traditionally, the model is trained to produce a latent language 
embedding for the given input as 
$\mathbf{z}_T = \Phi_{T}(\mathbf{x}_T;\texttt{CLS})$,
where `;' denotes concatenation and \texttt{CLS} is the special token for classification.
The latent embedding $\mathbf{z}_T$ can be used for downstream tasks.
In order to adapt the model on different tasks or different datasets,
one would have to finetune the large-scale pretrained model $\Phi_{T}$,
which is potentially very resource demanding and could be infeasible for models with billions of parameters.

As its name suggests, prompt learning provides
a set of learnable vectors $\mathbf{h}_T = [h_T^1, ..., h_T^k] $
called prompt vectors, that are fed to the pretrained model and
encourage it to produce desirable outputs. 
Formally, the prompt learning process can be written as
\begin{equation}
    \hat{\mathbf{z}}_T = \Phi_{T}(\mathbf{h}_T ;\mathbf{x}_T;\texttt{CLS}),
\end{equation}
Due to the flexibility of the prompt learning method,
one can adapt the model on new tasks or new datasets
by only training the lightweight prompt vectors $\mathbf{h}_T$,
rather than finetuning the heavy 
pretrained model $\Phi_{T}$. 
This method was originally proposed in the NLP community~\citep{LiLiang2021,Lester2021},
and it was later used to prompt the language branch in pretrained visual-language models~\cite{Ju2021,Zhou2021}.

Recently, the prompt learning technique has also been applied to 
large-scale pretrained \emph{visual} models. A pretrained visual model $\Phi_{V}(.)$, -- typically a variant of the Vision Transformer~\citep{Dosovitskiy2020}, is adapted using prompt vectors $\mathbf{h}_V$, yielding an image embedding $\hat{\mathbf{z}}_V$.
Within the previous works~\citep{Jia2022,Bahng2022} that apply prompt learning
in this way, %. Note that the prompt vectors $\mathbf{h}_V$ 
\citep{Jia2022} apply these prompts deep within the internal computations of transformers, whereas those in~\citep{Bahng2022} 
are in the pixel space.
\subsection{Prompt Generator Networks}
\label{method:pgn}
Although learning prompt vectors brings flexibility to the pretrained model,
a key limitation of the classic prompt learning method
is that the prompt vectors are shared within the dataset and task.
In other words, the prompt vectors are conditioned on the domain.
However,
what exactly constitutes a domain is somewhat arbitrary, 
\textit{e.g.},  ImageNet both contains fine-grained distinctions (such as 120 different dog breeds) and very coarse ones (such as a single mushroom class).
Therefore, having a single set of learned vectors for adaptation results in modeling capacity being wasted on items that might already be well encoded. 
In this section we introduce our Prompt Generation Network as a more flexible alternative.

\Cref{fig:method} shows an overview of our method.
Given the input image 
$I_i \in \mathbb{R}^{3\times H \times W}$ and the 
pretrained vision model $\Phi_{V}$,
we first cut the image into
$s\times s$ patches $\{c_1, c_2, ..., c_{H' \cdot W'}\}$,
where $c_j\in \mathbb{R}^{3\times s\times s}$, $H'=H/s$ and $W'=W/s$.
Then, we encode the patches with a linear layer $E(.)$.
To construct the visual inputs
$\mathbf{x}_V = [ E(c_1), E(c_2), ..., E(c_{H' \cdot W'}) ]$.
Rather than introducing a set of \emph{shared} prompt vectors as described in Section~\ref{method:vp},
we propose to use a set of \textit{input-dependent} prompt vectors.
Formally, we use a function $f(.)$ to learn the dependency
\begin{align}
    \hat{\mathbf{h}}^{i}_V &= f(I_i),
    % \hat{\mathbf{z}}^{i}_V &= \Phi_{V}(\mathbf{h}^{i}_V ;\mathbf{x}_V)
\end{align}
where the function $f(.)$ is learned by a neural network.
While it is possible to directly transform the inputs continuously to the prompt vectors, in practice this results in a high number of parameters. This is due to the size of the fully-connected layers, which is proportional to the large input dimensionalities of transformers. 

Instead, we propose to use a Token Library $\mathbf{L}\in \mathbb{R}^{N\times C}$ that consists of $N$ learnable feature vectors with channel dimension $C$.
Compared to works that use memory~\citep{sukhbaatar2015end,han2020memory,kumar2016ask}, 
the Token Library's purpose is not to learn characteristics of the dataset, 
but instead to steer the pretrained model towards a certain dataset,
therefore saving modeling capacity. 
The prompt generation network learns to generate $K$ prompt vectors
$\hat{\mathbf{h}}^i_V = [\hat{\mathbf{h}}^{i1}_V,...,\hat{\mathbf{h}}^{iK}_V]$,
each of which is a combination of the feature vectors from the Token Library $\mathbf{L}$. 
In detail,
\begin{align}
    \label{combination_equation}
    \hat{\mathbf{h}}^{i}_V &= f_{\theta,\mathbf{L}}(I_i) = 
    \text{Softmax}(g_{\theta}(I_i))\cdot \mathbf{L} \in{\mathbb{R}^{K\times C}},
\end{align}
where $g_{\theta}(.)$ is a learned mapping
$\{ \mathbb{R}^{3\times H\times W} \mapsto \mathbb{R}^{K\times N} \}$,
 using a lightweight neural network parameterized by $\theta$. Compared to some works in continual learning~\cite{wang2022learning, wang2022dualprompt}, the generated combination given by $g_{\theta}(.)$ is independent of the pre-trained representation given by $\Phi_{V}$, enabling the PGN to learn features that are optimized for its task of providing input-dependent prompts.
 
Finally all prompt vectors are fed into the frozen transformer encoder,
\begin{align}
    \label{visual_tfm}
    \hat{\mathbf{z}}^i_V &= \Phi_{V}(\hat{\mathbf{h}}^i_V ;\mathbf{x}^i_V; \texttt{CLS}).
\end{align}
\subsection{Prompt inversion for input-space prompting \label{method:pi}}
So far, the prompt vectors learned by a PGN were fed to the frozen pretrained model between the initial projection layer and the transformer blocks. 
Yet, if the first layer of the transformer cannot be modified, \textit{e.g.} because it is being accessed via an API or is implemented in hardware, this is problematic.
In order to solve this, we use a ``Prompt Inversion'' operation, 
a simple mechanism that transforms the learned prompts to input RGB patches, which can be simply attached to the bottom of the incoming image. This yields the same prompting capabilities, and is compatible with the recent emergence of inference APIs. In this setup, the PGN is first trained either in a local environment where the gradients of the frozen model are available, or remotely using an API that can return gradient information. Then, the trained PGN is deployed on the client-side and sends the adapted inputs to the remotely hosted model. 

\begin{figure} % 8 is relative width of caption compared to image
    % \centering
    % \begin{minipage}{.5\columnwidth}
        \centering
        \includegraphics[width=.35\textwidth]{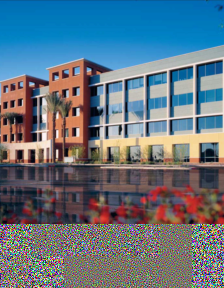}
    % \end{minipage}%
    % \hfill%
    % \begin{minipage}{.5\columnwidth}
    %     \begin{center}
    %         \includegraphics[width=.5\textwidth]{images/prompt_to_patch_2.png}
    %     \end{center}
    % \end{minipage}
    \caption{\textbf{Prompt Inversion for visual prompting.} At inference time, PGN prompts can be mapped to the input space to completely separate the adaptation step from the model's internal computation. In this instance, the PGN outputs 14 prompts which can be visualized as 2 rows containing 7 prompt-patches each. \label{fig:prompt_to_patch}}
\end{figure}

Prompt inversion leverages the linearity of the first patch-projection layer and enables fast training by learning directly in prompt-space, while at the same time allowing for \emph{input-only prompting} during inference and downstream applications.
In detail, we can transform a single prompt vector $\hat{\mathbf{h}}^{ij}_V$ into a patch by
\begin{equation}
    \mathcal{P}_{(\hat{\mathbf{h}}^{ij}_V)} = \big( \hat{\mathbf{h}}^{ij}_V - E_{\text{pos}} \big) W_{\text{proj}}^{-1},
\end{equation}
\noindent where $E_{\text{pos}}$ denotes the positional embedding and
$W_{\text{proj}}^{-1}$ is the pseudo-inverse of the linear projection weight matrix. Note that this matrix needs to be computed only once as it stays frozen.
The inverted prompt $\mathcal{P}_{(\hat{\mathbf{h}}^{ij}_V)} \in \mathbb{R}^{3 \times s \times s}$ is a RGB patch and can simply be appended to the input image and then served to the frozen transformer. 
In practice, the number of prompts is kept divisible by the patch size $s$ to ensures that the images retain a rectangular format. 
An example of the result of this process for a ViT-B/32 is shown in \cref{fig:prompt_to_patch} -- note that for ViTs with smaller patch-sizes the additional input patches occupy much less space. 
\section{Experiments}

\paragraph{Datasets.} 
In our experiments we cover 12 public image datasets:
CIFAR100 \& CIFAR10~\citep{krizhevsky2009cifar}, 
Oxford Flowers~\citep{nilsback2008flower},
\mbox{Food101~\citep{bossard2014food}},
EuroSAT~\citep{helber2019eurosat},
SUN397~\citep{xiao2010sun}, 
UCF101~\citep{soomro2012ucf101},
SVHN~\citep{netzer2011svhn}, 
Oxford-IIIT Pets~\citep{parkhi2012cats},
DTD~\citep{cimpoi2014describing}, 
RESISC~\citep{cheng2017resisc} 
and CLEVR~\citep{johnson2017clevr}.
Out of these, we chose CIFAR100 and SUN397 for ablations, as they vary in resolution, difficulty and the degree of spatial distribution of relevant image features.
Please refer to the supplementary materials for the statistics for each dataset.

\paragraph{Architectures.} 
We use the ViT-B/32 models and use CLIP pretraining~\citep{Radford2021} obtained weights as default but compare against supervised~\cite{Dosovitskiy2020} and DINO~\cite{caron2021emerging} self-supervised weights. 
The ViT weights are always kept frozen during our experiments.
In the PGN, 
we experiment with the lightweight ResNet-based architectures~\citep{he2016resnet},
namely ResNet10 and ResNet18. 
We obtain the ResNet10 architecture by 
reducing the layers at each residual block from ResNet18
and also reducing the depth of the first layer to only 16.
More details are provided supplementary materials.

\paragraph{Additional details and analyses.} We provide training details and additional experiments and analyses in the supplementary materials, including a qualitative analysis of the effect of PGNs on the computation within the Vision Transformer.

% \subsection{Proof of principle \label{s:exp-proof}}
\subsection{Motivational Analysis \label{s:exp-proof}}

\begin{figure}[t]
  \begin{minipage}[c]{0.4\columnwidth}
    \includegraphics[width=\textwidth]{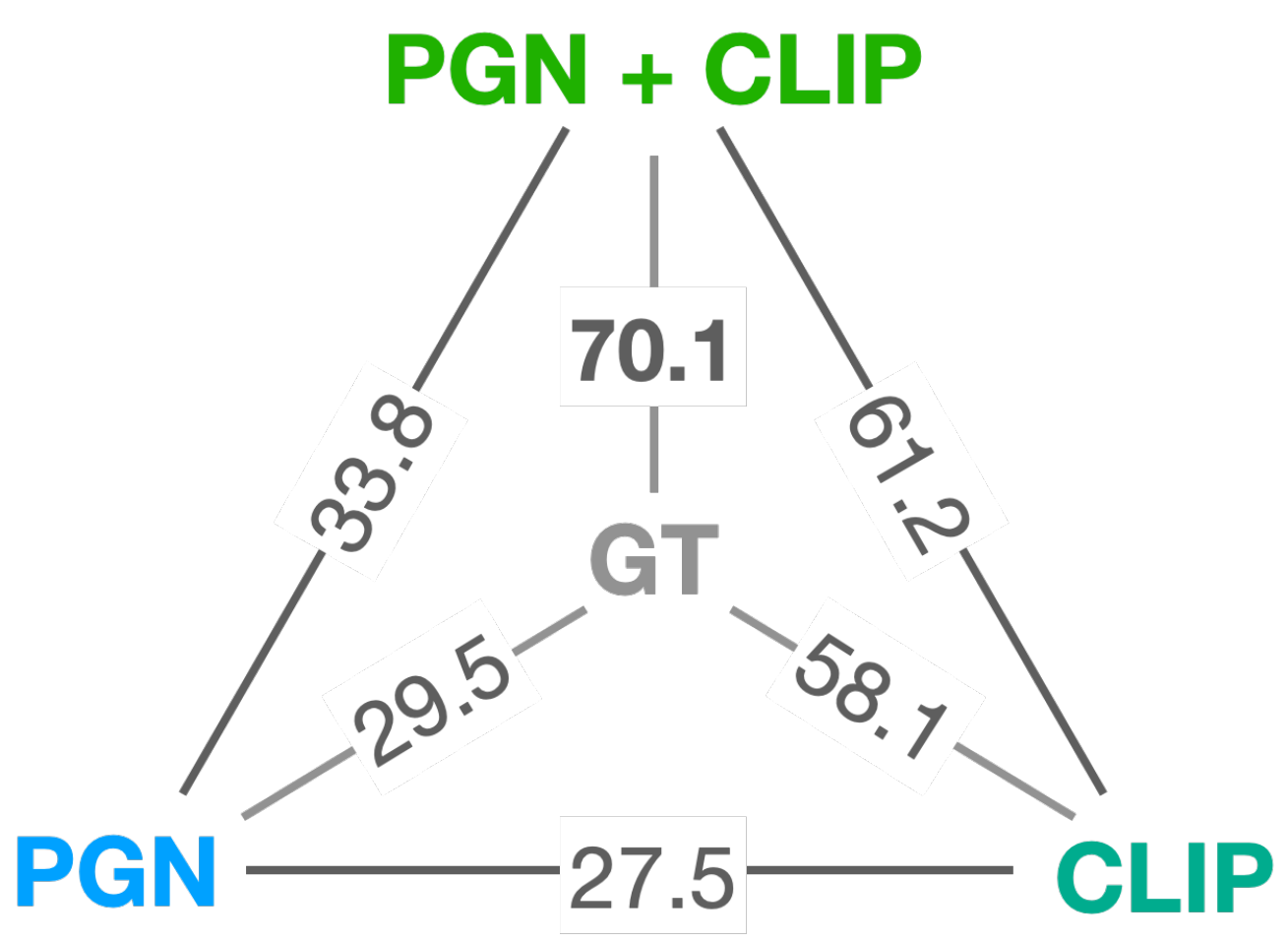}
    \caption{\textbf{Feature similarities.} We show the mutual information between the components and the ground-truth labels (details in the supplementary materials).
    PGN learns very different embeddings to CLIP but when combined achieves strong alignment with the ground-truth labels.
    \label{fig:nmis}}
  \end{minipage}\hspace{1em}
  \begin{minipage}[c]{0.55\columnwidth}
    \input{figures/parameter_efficiency_plot.tex}
  \end{minipage}
  % \vspace{-2em}
\end{figure}

In this section we investigate the basic feasibility of our proposed method for generating input-dependent prompts via PGNs.
For this, in~\cref{tab:proof_of_principle}, we compare the results of prompting a CLIP ViT-B/32 model to correctly select the right class in a ``zero-shot'' manner with text-embeddings automatically generated as ``This is a photo of a [class name]''. 
We observe that the zero-shot baseline of CLIP approximately matches that of training the network underlying the PGN ($g_{\theta}$ in \cref{fig:method}) in a standalone, supervised manner with accuracy around 63\%.
Despite this, the two models symbiotically increase the performance by +16\% when combined in our PGN method -- far outperforming its isolated variants.

\begin{table}[b]
    \centering
    % \begin{table}[t]
% \vspace{-2em}
    % \begin{center}
    
    \setlength\tabcolsep{0.3em}
    \begin{small}
    \begin{tabular}{ccc|c}
        \toprule
        &\color{gray}PGN backbone (alone) & CLIP & CLIP w/ PGN \\ 
        \midrule
        CIFAR-100 & \color{gray}63.7 & 63.1 & 79.3  \\
        \bottomrule
    \end{tabular}
    \end{small}
    \vspace{1em}
        % \vspace{-0.5em}
    % \end{center}
    % \vskip -0.2in
% \end{table}
    \caption{\textbf{Synergies.} We show performances of training our PGN's backbone in a supervised manner against CLIP ViT-B/32 in a zero-shot manner and when combined with PGN. \label{tab:proof_of_principle}}
\end{table}
This is further demonstrated in~\cref{fig:nmis}, where we compare the similarities of the representations of the various components. 
For this we cluster the outputs of each visual encoder (PGN, CLIP and CLIP+PGN) unsupervisedly and measure the pairwise alignment using normalised mutual information (see the supplementary materials for details). 
We find that PGN+CLIP's features are closest to the ground-truth (NMI of 70.1\%) compared to both PGN (29.5\%) and standalone CLIP (58.1\%), in line with later results of~\cref{large_comparison}. 
The low performance of only the PGN also demonstrates that it is not doing the ``heavy-lifting'', as its outputs are least aligned to the ground-truth labels.
Instead, PGN and CLIP's outputs share the lowest similarity of only 27.5\%. The superior performance of the combined model there results from the PGN learning features that are quite dissimilar and therefore more informative to the frozen CLIP model.

\subsection{Ablation studies \label{s:exp-ablation}}

\begin{table}
\begin{tabular}{ccc}
(a) Number of prompts.&(b) Token Library size&(c) Model architecture.\\
\bmvaHangBox{% \begin{table}[h]
    % \begin{center}
    \vspace{0.35em}
    \setlength\tabcolsep{0.2em}
    \begin{small}
    \begin{tabular}{lcc}
        \toprule
        Num. & C100 & SUN397 \\
        \midrule
        1 & 75.8 & 68.3 \\
        % 2 & 76.7 & 68.7 \\
        4 & 77.8 & 69.6 \\
        \cellcolor{gray!25} 8 & 77.9 &  \textbf{70.6} \\
        16 & \textbf{78.3} & 70.0 \\
        64 & 77.3 & 68.9 \\
        \bottomrule
    \end{tabular}
    \end{small}
    \label{abl_prompts}
    % \end{center}
% \end{table}
}&
\bmvaHangBox{% \begin{table}[h]
    % \begin{center}
    \vspace{0.35em}
    \label{abl_dict-size}
    \setlength\tabcolsep{0.2em}
    \begin{small}
    \begin{tabular}{lcc}
        \toprule
        Size & C100 & SUN397 \\
        \midrule
        8 & 75.3 & 69.0 \\
        16 & 76.4 & 69.6 \\
        \cellcolor{gray!25}64 & 77.9 & \textbf{70.6}\\
        256 & 78.4 & \textbf{70.6}\\
        1024 & \textbf{78.6} & 69.9\\
        \bottomrule
    \end{tabular}
    \end{small}
    % \caption{Size of the token library.}
    % \end{center}
% \end{table}
}&
\bmvaHangBox{% \begin{table}[h]
    % \begin{center}
    % \captionof{subtable}{PGN backbone.}
    % \vspace{0.35em}
    \setlength\tabcolsep{0.3em}
    \begin{small}
    \begin{tabular}{lcc}
        \toprule
        Model (resolution) & C100 & SUN397 \\
        \midrule
        IIP & 72.1 & 69.6 \\
        MLP (10$^2$)& 73.5 & 69.6\\
        \cellcolor{gray!25}ResNet10 (64$^2$) & 77.9 & \textbf{70.6} \\
        ResNet10 (224$^2$) & \textbf{79.3} & \textbf{70.6} \\
        ResNet18 (224$^2$) & \textbf{79.3} & 69.6\\
        \bottomrule
    \end{tabular}
    \label{abl_architecture}
    \end{small}
    % \end{center}
% \end{table}
}
\vspace{0.4em}
\end{tabular}
\captionof{table}{\textbf{Ablations}. The configurations are evaluated on CIFAR100 (C100) and SUN397. We vary the
number of prompts provided to the frozen model, the size of the Token Library and the PGN
backbone. Default settings are in gray.}
\label{tab:ablations}
\end{table}

Next, we ablate and analyze the various choices of the PGN architecture and setting.
We conduct ablation studies on CIFAR100 and SUN397 in~\cref{tab:ablations}.
Unless otherwise stated, we learn $8$ prompts and a Token Library of size $64$ to speed up computations.

\paragraph{Number of prompts.}
In~\cref{tab:ablations}(a), we vary the number of prompts that the PGN generates.
We find that increasing the number of prompts generally yields a better performance until around $4$ prompts. 
This shows that even a modest number of 4-8 additional tokens -- when added to the 49 spatial plus 1 \texttt{CLS} token -- can yield significant benefits.
% \HTD{what are the 50 existing tokens?}

\paragraph{Token Library size.}
In~\cref{tab:ablations}(b), we compare the size of the Token Library's number of items. 
We find that larger Token Library generally leads to better performance, though no further improvement is observed beyond $256$ tokens. 
We conjecture the additional library items
beyond $256$ tokens only learn redundant information and
therefore do not improve.

\paragraph{Architectures.}
In~\cref{tab:ablations}(c), we ablate the backbone used for the PGN with input-independent prompts (IIP), a setup in which the prompts consists of learnable vectors that are constant for each input image.
With this as a baseline, we compare against input-dependent prompting methods, including a 2-layer MLP operating on the center $10\times 10$ pixels of an image, a ResNet10 at resolutions of $64\times 64$ and $224\times 224$, as well as a ResNet18 at resolutions of $224\times 224$.
First, we observe that the 2-layer MLP, a simple model with limited image modelling capability, obtains a small performance benefit over the IIP setup.
This might be explained by IIP being a strict subset of the input-dependent methods, as these could zero-out the input and instead supply constant prompts. 
The benefits of input-dependency is further demonstrated by using convolutional neural networks as backbones with gains of up to +5.8\% for CIFAR100 (77.9 vs. 72.1). 
Increasing input resolution from 64 to 224 also slightly improves (+1.4\%) accuracy for CIFAR100.
Finally, we observe that the overall performance saturates even when using a ResNet18 compared to the ResNet10, suggesting that the PGN can be lightweight and that the method automatically divides the labor between fine-grained recognition and providing prompts as cues.

\paragraph{Token Library vs direct learning.}
In~\cref{fig:parameter_efficiency} we study the approach of obtaining the prompts.
We compare our token library (TL) with 
the more direct approach that obtains the prompts through a linear transformation of the image features.
While we observe that both methods can be made to achieve similar performances, \cref{fig:parameter_efficiency} clearly demonstrates the superiority of the TL in terms of parameter efficiency.

\paragraph{Scaling to different frozen transformer models.}
Finally, we explore the use of different pretrained ViT backbones, from supervised training~\citep{Dosovitskiy2020} and from self-supervised training using DINO~\citep{caron2021emerging}.
First, we find that generally the performances are lower compared to those obtained from the CLIP backbone in~\cref{large_comparison}.
This means that adapting these models is more challenging, potentially owing to the vastly larger pretraining dataset of CLIP.
Second, we find that both IIP and PGN struggle with adapting the backbones to the SUN397 dataset.
This might be explained by: (i) the difference in image contents -- while SUN contains scene images, the pretraining datasets of the supervised ViT and DINO are strongly object-centric~\citep{kuznetsova2020open} -- and (ii) the relatively small number of images per class ($\approx$190).
Third, we find that, as in the case of CLIP, the PGN approach vastly outperforms the baseline IIP approach in adapting these models to the datasets, \textit{e.g.}, showing gains of 40\% for CIFAR100. 

\subsection{Large-scale comparisons \label{s:exp-large-scale}}
\begin{table*}[!htb]

\setlength\tabcolsep{0.15em}
\begin{center}
\begin{small}
\begin{tabular}{l|cccccccccccc|ccc}
\toprule
 Method & \scriptsize C100 & \scriptsize C10 & \scriptsize Flwrs & \scriptsize Food & \scriptsize EuSAT & \scriptsize SUN & \scriptsize UCF & \scriptsize SVHN & \scriptsize Pets & \scriptsize DTD & \scriptsize RESISC & \scriptsize CLEVR & Avg.  & $\Sigma$\,params \\
\midrule
CLIP+TP & 63.1 & 89.0 & 61.9 & 79.8 & 40.0 & 60.0 & 59.9 & 5.1 & 85.9 & 43.0 & 42.4 & 20.2 & 54.2 & - \\
% \midrule
\,\,+ VP & 75.3 & 94.2 & 70.3 & 78.9 & 96.4 & 60.6 & 66.1 & 88.4 & 85.0 & 57.1 & 84.5 & 81.4 & 78.2 & 0.94M \\ %70K *12
\midrule 
\textbf{\,\,+ PGN (ours)} & 79.3 & \cellcolor{green!25}\textbf{96.1} & 94.0 & \cellcolor{green!25}82.5 & \textbf{98.0} & 7\cellcolor{green!25}0.9 & 77.6 & 94.2 & \cellcolor{green!25}\textbf{91.5} & 71.5 & 92.1 & \cellcolor{green!25}\textbf{95.1} & \cellcolor{green!25}\textbf{86.9} & 12.4M \\
\midrule 
Linear ft. & {80.0} & 95.0 & 96.9 & \textbf{84.6} & 95.3 & \textbf{75.0} & \textbf{83.3} & 65.4 & 89.2 & \textbf{74.6} & {92.3} & 66.0 & 83.1 & 0.75M \\% \\ 740K\\
full-ft. & \textbf{82.1} & 95.8 & \textbf{97.4} & 80.5 & 97.9 & 64.0 & 80.9 & \textbf{95.7} & 88.5 & 72.3 & \textbf{93.3} & 94.4 & \textbf{86.9} & 1032M\\
% CLIP & FT & 82.1 & 95.8 & 97.4 & 80.5 & 97.9 & 64.0 & 80.9 & 95.7 & 88.5 & 72.3 & 93.3 & 94.4 & 86.9\\
\bottomrule
\end{tabular}
\end{small}
\vspace{0.6em}
\caption{\textbf{Performance across 12 datasets.} We compare using CLIP with text prompting (TP), visual prompting (VP) ~\cite{Bahng2022}, linear or full-finetuning (ft.) and our PGN method. 
The \colorbox{green!25}{green} shade indicates where PGN outperforms or matches full-finetuning performance. The bolding refers to best performance per dataset. 
We report the additional number of parameters of the visual encoder for each configuration. \label{large_comparison}}
\end{center}
\end{table*}
% linear layer: 768*(100+10+102+101+10+397+101+10+37+47+45+8) = 743424

Next, we compare PGNs to other input-based adaptation techniques on a wide range of vision datasets. 
Based on the findings of our ablations, we choose the ResNet10 as the PGN's backbone, outputting $16$ prompts from a Token Library of size $256$, feeding into a pretrained CLIP ViT-B/32. 
The results are shown in~\cref{large_comparison} and are directly comparable to the concurrent work of~Bahng \etal~\cite{Bahng2022}  on visual prompts (VP), from which we adapted the results of the first two rows.
From~\cref{large_comparison}, we first find that linear and full-finetuning and our method achieve the best performances on 4 out of 12 datasets each. 
However, when comparing the overall averaged accuracies, we find that our method achieves 86.9\%, matching the performance of the full-finetuning adaptation, and surpassing both VP and linear finetuning by 8\% and 3\%, respectively.
In the last column of~\cref{large_comparison}, we show the number of additional parameters for adapting to each dataset. 
We find that our PGN method requires almost two orders of magnitude fewer parameters than full-finetuning, while retaining the same overall performance. 

\begin{table}[!t]
  \begin{minipage}[c]{0.4\columnwidth}
    \includegraphics[width=.6\columnwidth]{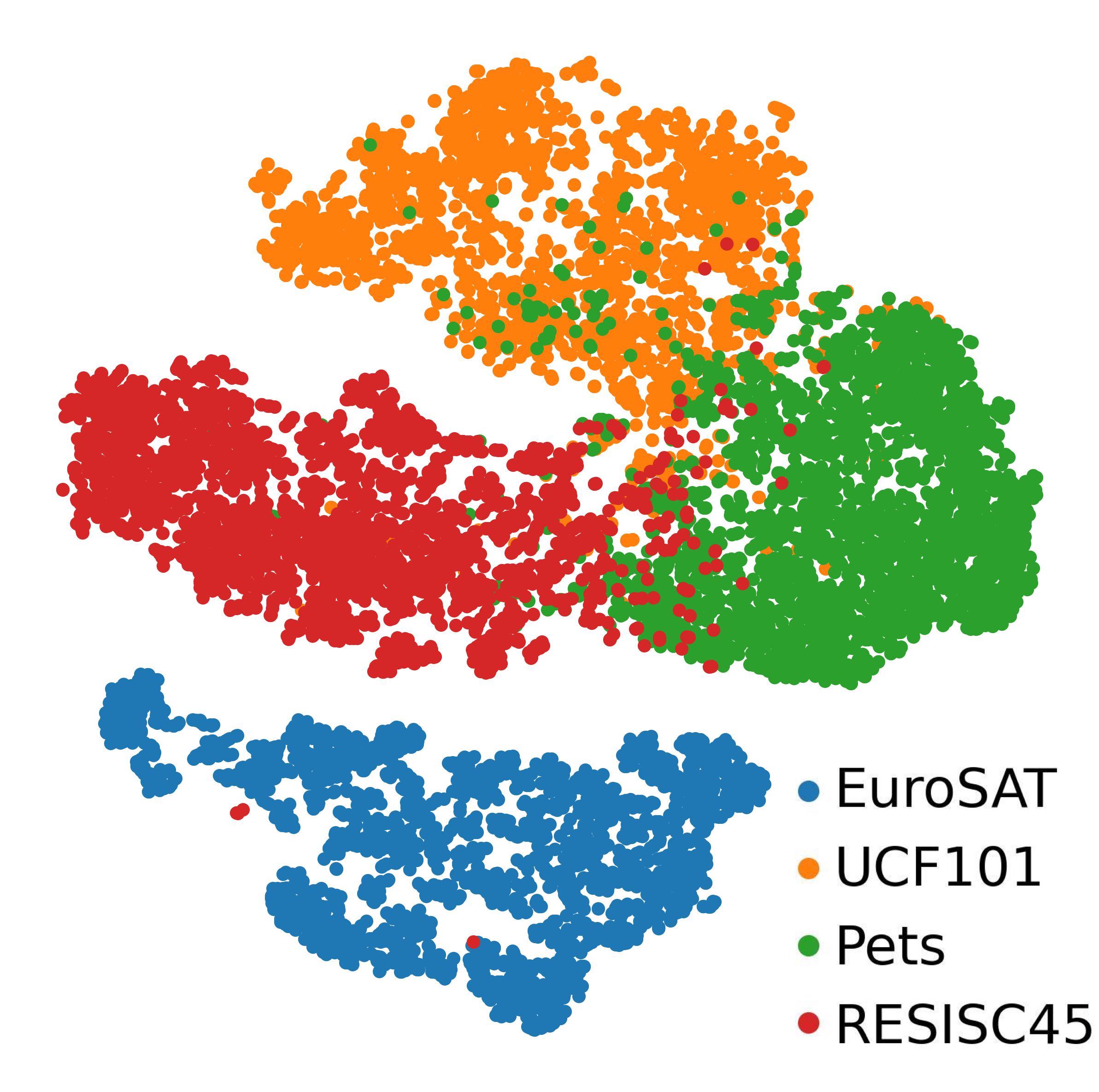}
    \captionof{figure}{\textbf{Automatic domain discovery.} \mbox{$t$-SNE}     visualisation of PGN outputs. PGN trained on a mixture of datasets automatically allocates the tokens in a manner that recovers the individual training domains.\label{fig:tsne}}
  \end{minipage}\hfill%
  \begin{minipage}[c]{0.55\columnwidth}
    \vspace{1em}
    % \begin{table}[h]
%     \begin{center}
%     \captionof{table}{\textbf{PGNs for different pretrained models.}}
%     \label{tab:dino-vit}
%     \setlength\tabcolsep{0.2em}
%     \begin{small}
%     \begin{tabular}{lc cc}
%         \toprule
%         & & CIFAR & SUN \\
%         \midrule
%         \multirow{2}{*}{DINO} & IIP \\
%         & PGN \\
%         \multirow{2}{*}{ViT} & IIP \\
%         & PGN \\
%         \bottomrule
%     \end{tabular}
%     \end{small}
%     \end{center}
%     \vskip -0.1in
% \end{table}

% \begin{table}[h]
    % \begin{center}
    \small
    \label{tab:dino-vit}
    \setlength\tabcolsep{0.8em}
    % \begin{small}
    \begin{tabular}{lcc c cc}
        \toprule
         & \multicolumn{2}{c}{DINO} &&  \multicolumn{2}{c}{ViT sup.}\\
         \cmidrule{2-3} \cmidrule{5-6}
         & IIP & PGN && IIP & PGN \\
        \midrule
        CIFAR100 & 13.2 & \textbf{53.0} && 18.8 & \textbf{50.7} \\
        SUN397 & 3.0 & \textbf{19.2} && 2.6 & \textbf{6.3} \\
        EuroSAT & 85.8 & \textbf{94.6} && 91.9 & \textbf{96.1} \\
        \bottomrule
    \end{tabular}
    % \end{small}
    % \end{center}
    
% \end{table}
    \vspace{0.6em}
    \caption{\textbf{Different pretrainings.} We compare PGN on two other ViT-B/32 backbones from self-supervised DINO~\citep{caron2021emerging} and from supervised training~\cite{Dosovitskiy2020}. Best results bolded.}
  \end{minipage}
\end{table}

\subsection{Efficient multi-dataset PGN}
Encouraged by the comparisons of the previous section, we investigate whether PGNs can be made even more efficient, by training a PGN on multiple datasets at the same time. The numerical results are shown in the supplementary materials. 

In~\cref{fig:tsne} we show a $t$-SNE visualisation of PGN outputs for four datasets. The PGN was trained in the PGN+CLIP setup of \cref{s:exp-large-scale}, on the union of the four datasets.
First, we find that the PGN learns features that are well-correlated with the domains, despite not having access to the individual item's dataset origin.
This means that such a procedure could be used for adapting to mixed domain or unsorted datasets.
We also observe that some domains contain images that have similar PGN outputs, notably UCF and Pets. 
This discovery might be explained by overlaps in depicted objects, \textit{e.g.}, UCF contains classes such as ``walking a dog'', while Pets contains multiple dog classes. 
\section{Discussion and conclusion}
\paragraph{Impact.}
The increasing size of large-scale pretrained models have had a centralizing effect: only large companies and institutions have the resources to collect data and train on the required scale. One recently explored option to share the benefits of these models is to make them available for experimentation and inference through APIs. 
In this work, we aim to empower future users of foundation model APIs by addressing the key limitation of the restricted adaptability of models in this setting. 
However, we also recognize that our adaptation method could potentially be used by malicious actors for unintended purposes and vigilant monitoring of usages will remain important.
\paragraph{Conclusion.} We propose the Prompt Generation Network (PGN), a simple and effective framework for learning input-dependent visual prompts.
Our method is parameter-efficient by leveraging a Token Library from which tokens are combined to yield the prompts using a lightweight network. Furthermore, it is compatible with recent developments that restrict access to model internals at inference, such as APIs or implementations on edge devices that require specialized hardware.
We demonstrate that PGN can be used to adapt CLIP, surpassing previous methods and even matching and exceeding the results of full-finetuning of the visual encoder, despite requiring two orders of magnitude less number of adapted weights.
Finally, we have demonstrated that PGN can be a scalable method for generic adaptation of frozen visual transformers by training them with a mixture of datasets.
%-------------------------------------------------------------------------

\bibliography{references.bib}

\begin{thebibliography}{50}
\providecommand{\natexlab}[1]{#1}
\providecommand{\url}[1]{\texttt{#1}}
\expandafter\ifx\csname urlstyle\endcsname\relax
  \providecommand{\doi}[1]{doi: #1}\else
  \providecommand{\doi}{doi: \begingroup \urlstyle{rm}\Url}\fi

\bibitem[Bahng et~al.(2022)Bahng, Jahanian, Sankaranarayanan, and Isola]{Bahng2022}
Hyojin Bahng, Ali Jahanian, Swami Sankaranarayanan, and Phillip Isola.
\newblock Exploring visual prompts for adapting large-scale models.
\newblock \emph{arXiv:2203.17274}, 2022.

\bibitem[Bossard et~al.(2014)Bossard, Guillaumin, and Van~Gool]{bossard2014food}
Lukas Bossard, Matthieu Guillaumin, and Luc Van~Gool.
\newblock Food-101 -- mining discriminative components with random forests.
\newblock In \emph{ECCV}, 2014.

\bibitem[Brown et~al.(2020)Brown, Mann, Ryder, Subbiah, Kaplan, Dhariwal, Neelakantan, Shyam, Sastry, Askell, et~al.]{Brown2020}
Tom Brown, Benjamin Mann, Nick Ryder, Melanie Subbiah, Jared~D Kaplan, Prafulla Dhariwal, Arvind Neelakantan, Pranav Shyam, Girish Sastry, Amanda Askell, et~al.
\newblock Language models are few-shot learners.
\newblock \emph{NeurIPS}, 33:\penalty0 1877--1901, 2020.

\bibitem[Cai et~al.(2020)Cai, Gan, Zhu, and Han]{Han2020tiny}
Han Cai, Chuang Gan, Ligeng Zhu, and Song Han.
\newblock Tinytl: Reduce memory, not parameters for efficient on-device learning.
\newblock In \emph{NeurIPS}, pages 11285--11297, 2020.

\bibitem[Caron et~al.(2021)Caron, Touvron, Misra, J{\'e}gou, Mairal, Bojanowski, and Joulin]{caron2021emerging}
Mathilde Caron, Hugo Touvron, Ishan Misra, Herv{\'e} J{\'e}gou, Julien Mairal, Piotr Bojanowski, and Armand Joulin.
\newblock Emerging properties in self-supervised vision transformers.
\newblock In \emph{ICCV}, pages 9650--9660, 2021.

\bibitem[Cheng et~al.(2017)Cheng, Han, and Lu]{cheng2017resisc}
Gong Cheng, Junwei Han, and Xiaoqiang Lu.
\newblock Remote sensing image scene classification: Benchmark and state of the art.
\newblock \emph{Proceedings of the IEEE}, 105\penalty0 (10):\penalty0 1865--1883, 2017.

\bibitem[Cimpoi et~al.(2014)Cimpoi, Maji, Kokkinos, Mohamed, and Vedaldi]{cimpoi2014describing}
Mircea Cimpoi, Subhransu Maji, Iasonas Kokkinos, Sammy Mohamed, and Andrea Vedaldi.
\newblock Describing textures in the wild.
\newblock In \emph{CVPR}, pages 3606--3613, 2014.

\bibitem[Dehghani et~al.(2023)Dehghani, Djolonga, Mustafa, Padlewski, Heek, Gilmer, Steiner, Caron, Geirhos, Alabdulmohsin, et~al.]{dehghani2023scaling}
Mostafa Dehghani, Josip Djolonga, Basil Mustafa, Piotr Padlewski, Jonathan Heek, Justin Gilmer, Andreas Steiner, Mathilde Caron, Robert Geirhos, Ibrahim Alabdulmohsin, et~al.
\newblock Scaling vision transformers to 22 billion parameters.
\newblock \emph{arXiv:2302.05442}, 2023.

\bibitem[Deng et~al.(2009)Deng, Dong, Socher, Li, Li, and Fei-Fei]{Deng2009}
Jia Deng, Wei Dong, Richard Socher, Li-Jia Li, Kai Li, and Li~Fei-Fei.
\newblock Imagenet: A large-scale hierarchical image database.
\newblock In \emph{2009 IEEE Conference on Computer Vision and Pattern Recognition}, pages 248--255, 2009.
\newblock \doi{10.1109/CVPR.2009.5206848}.

\bibitem[Dosovitskiy et~al.(2020)Dosovitskiy, Beyer, Kolesnikov, Weissenborn, Zhai, Unterthiner, Dehghani, Minderer, Heigold, Gelly, Uszkoreit, and Houlsby]{Dosovitskiy2020}
Alexey Dosovitskiy, Lucas Beyer, Alexander Kolesnikov, Dirk Weissenborn, Xiaohua Zhai, Thomas Unterthiner, Mostafa Dehghani, Matthias Minderer, Georg Heigold, Sylvain Gelly, Jakob Uszkoreit, and Neil Houlsby.
\newblock An image is worth 16x16 words: Transformers for image recognition at scale.
\newblock \emph{ICLR}, 2020.

\bibitem[Dubey et~al.(2024)Dubey, Jauhri, Pandey, Kadian, Al-Dahle, Letman, Mathur, Schelten, Yang, Fan, et~al.]{dubey2024llama}
Abhimanyu Dubey, Abhinav Jauhri, Abhinav Pandey, Abhishek Kadian, Ahmad Al-Dahle, Aiesha Letman, Akhil Mathur, Alan Schelten, Amy Yang, Angela Fan, et~al.
\newblock The llama 3 herd of models.
\newblock \emph{arXiv preprint arXiv:2407.21783}, 2024.

\bibitem[Elsayed et~al.(2018)Elsayed, Goodfellow, and Sohl-Dickstein]{elsayed2018adversarial}
Gamaleldin~F Elsayed, Ian Goodfellow, and Jascha Sohl-Dickstein.
\newblock Adversarial reprogramming of neural networks.
\newblock \emph{arXiv preprint arXiv:1806.11146}, 2018.

\bibitem[Gao et~al.(2021)Gao, Geng, Zhang, Ma, Fang, Zhang, Li, and Qiao]{gao2021clip}
Peng Gao, Shijie Geng, Renrui Zhang, Teli Ma, Rongyao Fang, Yongfeng Zhang, Hongsheng Li, and Yu~Qiao.
\newblock Clip-adapter: Better vision-language models with feature adapters.
\newblock \emph{arXiv:2110.04544}, 2021.

\bibitem[Goyal et~al.(2021)Goyal, Caron, Lefaudeux, Xu, Wang, Pai, Singh, Liptchinsky, Misra, Joulin, et~al.]{goyal2021self}
Priya Goyal, Mathilde Caron, Benjamin Lefaudeux, Min Xu, Pengchao Wang, Vivek Pai, Mannat Singh, Vitaliy Liptchinsky, Ishan Misra, Armand Joulin, et~al.
\newblock Self-supervised pretraining of visual features in the wild.
\newblock \emph{arXiv:2103.01988}, 2021.

\bibitem[Han et~al.(2020)Han, Xie, and Zisserman]{han2020memory}
Tengda Han, Weidi Xie, and Andrew Zisserman.
\newblock Memory-augmented dense predictive coding for video representation learning.
\newblock In \emph{ECCV}, pages 312--329. Springer, 2020.

\bibitem[He et~al.(2016)He, Zhang, Ren, and Sun]{he2016resnet}
Kaiming He, Xiangyu Zhang, Shaoqing Ren, and Jian Sun.
\newblock Deep residual learning for image recognition.
\newblock In \emph{CVPR}, pages 770--778, 2016.

\bibitem[Helber et~al.(2019)Helber, Bischke, Dengel, and Borth]{helber2019eurosat}
Patrick Helber, Benjamin Bischke, Andreas Dengel, and Damian Borth.
\newblock Eurosat: A novel dataset and deep learning benchmark for land use and land cover classification.
\newblock \emph{IEEE Journal of Selected Topics in Applied Earth Observations and Remote Sensing}, 2019.

\bibitem[Hendrycks et~al.(2019)Hendrycks, Zhao, Basart, Steinhardt, and Song]{hendrycks2019nae}
Dan Hendrycks, Kevin Zhao, Steven Basart, Jacob Steinhardt, and Dawn Song.
\newblock Natural adversarial examples.
\newblock \emph{arXiv preprint arXiv:1907.07174}, 2019.

\bibitem[Hendrycks et~al.(2021)Hendrycks, Basart, Mu, Kadavath, Wang, Dorundo, Desai, Zhu, Parajuli, Guo, Song, Steinhardt, and Gilmer]{hendrycks2021many}
Dan Hendrycks, Steven Basart, Norman Mu, Saurav Kadavath, Frank Wang, Evan Dorundo, Rahul Desai, Tyler Zhu, Samyak Parajuli, Mike Guo, Dawn Song, Jacob Steinhardt, and Justin Gilmer.
\newblock The many faces of robustness: A critical analysis of out-of-distribution generalization.
\newblock \emph{ICCV}, 2021.

\bibitem[Houlsby et~al.(2019)Houlsby, Giurgiu, Jastrzebski, Morrone, De~Laroussilhe, Gesmundo, Attariyan, and Gelly]{houlsby2019parameter}
Neil Houlsby, Andrei Giurgiu, Stanislaw Jastrzebski, Bruna Morrone, Quentin De~Laroussilhe, Andrea Gesmundo, Mona Attariyan, and Sylvain Gelly.
\newblock Parameter-efficient transfer learning for nlp.
\newblock \emph{ICML}, 2019.

\bibitem[Jia et~al.(2021)Jia, Yang, Xia, Chen, Parekh, Pham, Le, Sung, Li, and Duerig]{Jia2021}
Chao Jia, Yinfei Yang, Ye~Xia, Yi-Ting Chen, Zarana Parekh, Hieu Pham, Quoc Le, Yun-Hsuan Sung, Zhen Li, and Tom Duerig.
\newblock Scaling up visual and vision-language representation learning with noisy text supervision.
\newblock In \emph{ICML}, pages 4904--4916. PMLR, 2021.

\bibitem[Jia et~al.(2022)Jia, Tang, Chen, Cardie, Belongie, Hariharan, and Lim]{Jia2022}
Menglin Jia, Luming Tang, Bor-Chun Chen, Claire Cardie, Serge Belongie, Bharath Hariharan, and Ser-Nam Lim.
\newblock Visual prompt tuning.
\newblock \emph{ECCV}, 2022.

\bibitem[Johnson et~al.(2017)Johnson, Hariharan, Van Der~Maaten, Fei-Fei, Lawrence~Zitnick, and Girshick]{johnson2017clevr}
Justin Johnson, Bharath Hariharan, Laurens Van Der~Maaten, Li~Fei-Fei, C~Lawrence~Zitnick, and Ross Girshick.
\newblock Clevr: A diagnostic dataset for compositional language and elementary visual reasoning.
\newblock In \emph{CVPR}, pages 2901--2910, 2017.

\bibitem[Ju et~al.(2022)Ju, Han, Zheng, Zhang, and Xie]{Ju2021}
Chen Ju, Tengda Han, Kunhao Zheng, Ya~Zhang, and Weidi Xie.
\newblock Prompting visual-language models for efficient video understanding.
\newblock In \emph{ECCV}, 2022.

\bibitem[Kloberdanz et~al.(2021)Kloberdanz, Tian, and Le]{kloberdanz2021improved}
Eliska Kloberdanz, Jin Tian, and Wei Le.
\newblock An improved (adversarial) reprogramming technique for neural networks.
\newblock In \emph{International Conference on Artificial Neural Networks}, pages 3--15. Springer, 2021.

\bibitem[Krizhevsky et~al.(2009)Krizhevsky, Hinton, et~al.]{krizhevsky2009cifar}
Alex Krizhevsky, Geoffrey Hinton, et~al.
\newblock Learning multiple layers of features from tiny images.
\newblock \emph{Technical Report}, 2009.

\bibitem[Kumar et~al.(2016)Kumar, Irsoy, Ondruska, Iyyer, Bradbury, Gulrajani, Zhong, Paulus, and Socher]{kumar2016ask}
Ankit Kumar, Ozan Irsoy, Peter Ondruska, Mohit Iyyer, James Bradbury, Ishaan Gulrajani, Victor Zhong, Romain Paulus, and Richard Socher.
\newblock Ask me anything: Dynamic memory networks for natural language processing.
\newblock In \emph{ICML}, pages 1378--1387. PMLR, 2016.

\bibitem[Kuznetsova et~al.(2020)Kuznetsova, Rom, Alldrin, Uijlings, Krasin, Pont-Tuset, Kamali, Popov, Malloci, Kolesnikov, et~al.]{kuznetsova2020open}
Alina Kuznetsova, Hassan Rom, Neil Alldrin, Jasper Uijlings, Ivan Krasin, Jordi Pont-Tuset, Shahab Kamali, Stefan Popov, Matteo Malloci, Alexander Kolesnikov, et~al.
\newblock The open images dataset v4.
\newblock \emph{IJCV}, pages 1956--1981, 2020.

\bibitem[Lester et~al.(2021)Lester, Al-Rfou, and Constant]{Lester2021}
Brian Lester, Rami Al-Rfou, and Noah Constant.
\newblock The power of scale for parameter-efficient prompt tuning.
\newblock \emph{EMNLP}, 2021.

\bibitem[Li and Liang(2021)]{LiLiang2021}
Xiang~Lisa Li and Percy Liang.
\newblock Prefix-tuning: Optimizing continuous prompts for generation.
\newblock \emph{ACL}, 2021.

\bibitem[Netzer et~al.(2011)Netzer, Wang, Coates, Bissacco, Wu, and Ng]{netzer2011svhn}
Yuval Netzer, Tao Wang, Adam Coates, Alessandro Bissacco, Bo~Wu, and Andrew~Y Ng.
\newblock Reading digits in natural images with unsupervised feature learning.
\newblock \emph{NeurIPS Workshop on Deep Learning and Unsupervised Feature Learning}, 2011.

\bibitem[Nilsback and Zisserman(2008)]{nilsback2008flower}
Maria-Elena Nilsback and Andrew Zisserman.
\newblock Automated flower classification over a large number of classes.
\newblock In \emph{Sixth Indian Conference on Computer Vision, Graphics \& Image Processing}, pages 722--729. IEEE, 2008.

\bibitem[Parkhi et~al.(2012)Parkhi, Vedaldi, Zisserman, and Jawahar]{parkhi2012cats}
Omkar~M Parkhi, Andrea Vedaldi, Andrew Zisserman, and CV~Jawahar.
\newblock Cats and dogs.
\newblock In \emph{CVPR}, pages 3498--3505. IEEE, 2012.

\bibitem[Pfeiffer et~al.(2020)Pfeiffer, R{\"u}ckl{\'e}, Poth, Kamath, Vuli{\'c}, Ruder, Cho, and Gurevych]{pfeiffer2020adapterhub}
Jonas Pfeiffer, Andreas R{\"u}ckl{\'e}, Clifton Poth, Aishwarya Kamath, Ivan Vuli{\'c}, Sebastian Ruder, Kyunghyun Cho, and Iryna Gurevych.
\newblock Adapterhub: A framework for adapting transformers.
\newblock \emph{EMNLP}, 2020.

\bibitem[Radford et~al.(2019)Radford, Wu, Child, Luan, Amodei, and Sutskever]{Radford2019}
Alec Radford, Jeffrey Wu, Rewon Child, David Luan, Dario Amodei, and Ilya Sutskever.
\newblock Language models are unsupervised multitask learners.
\newblock \emph{OpenAI Technical Report}, 2019.

\bibitem[Radford et~al.(2021)Radford, Kim, Hallacy, Ramesh, Goh, Agarwal, Sastry, Askell, Mishkin, Clark, Krueger, and Sutskever]{Radford2021}
Alec Radford, Jong~Wook Kim, Chris Hallacy, Aditya Ramesh, Gabriel Goh, Sandhini Agarwal, Girish Sastry, Amanda Askell, Pamela Mishkin, Jack Clark, Gretchen Krueger, and Ilya Sutskever.
\newblock Learning transferable visual models from natural language supervision.
\newblock \emph{ICML}, 2021.

\bibitem[Recht et~al.(2019)Recht, Roelofs, Schmidt, and Shankar]{recht2019imagenet}
Benjamin Recht, Rebecca Roelofs, Ludwig Schmidt, and Vaishaal Shankar.
\newblock Do imagenet classifiers generalize to imagenet?
\newblock In \emph{International conference on machine learning}, pages 5389--5400. PMLR, 2019.

\bibitem[Soomro et~al.(2012)Soomro, Zamir, and Shah]{soomro2012ucf101}
Khurram Soomro, Amir~Roshan Zamir, and Mubarak Shah.
\newblock Ucf101: A dataset of 101 human actions classes from videos in the wild.
\newblock \emph{arXiv preprint arXiv:1212.0402}, 2012.

\bibitem[Sukhbaatar et~al.(2015)Sukhbaatar, Weston, Fergus, et~al.]{sukhbaatar2015end}
Sainbayar Sukhbaatar, Jason Weston, Rob Fergus, et~al.
\newblock End-to-end memory networks.
\newblock \emph{NeurIPS}, 28, 2015.

\bibitem[Wang et~al.(2019)Wang, Ge, Lipton, and Xing]{wang2019learning}
Haohan Wang, Songwei Ge, Zachary Lipton, and Eric~P Xing.
\newblock Learning robust global representations by penalizing local predictive power.
\newblock In \emph{Advances in Neural Information Processing Systems}, pages 10506--10518, 2019.

\bibitem[Wang et~al.(2022{\natexlab{a}})Wang, Zhang, Ebrahimi, Sun, Zhang, Lee, Ren, Su, Perot, Dy, et~al.]{wang2022dualprompt}
Zifeng Wang, Zizhao Zhang, Sayna Ebrahimi, Ruoxi Sun, Han Zhang, Chen-Yu Lee, Xiaoqi Ren, Guolong Su, Vincent Perot, Jennifer Dy, et~al.
\newblock Dualprompt: Complementary prompting for rehearsal-free continual learning.
\newblock In \emph{Computer Vision--ECCV 2022: 17th European Conference, Tel Aviv, Israel, October 23--27, 2022, Proceedings, Part XXVI}, pages 631--648. Springer, 2022{\natexlab{a}}.

\bibitem[Wang et~al.(2022{\natexlab{b}})Wang, Zhang, Lee, Zhang, Sun, Ren, Su, Perot, Dy, and Pfister]{wang2022learning}
Zifeng Wang, Zizhao Zhang, Chen-Yu Lee, Han Zhang, Ruoxi Sun, Xiaoqi Ren, Guolong Su, Vincent Perot, Jennifer Dy, and Tomas Pfister.
\newblock Learning to prompt for continual learning.
\newblock In \emph{Proceedings of the IEEE/CVF Conference on Computer Vision and Pattern Recognition}, pages 139--149, 2022{\natexlab{b}}.

\bibitem[Wortsman et~al.(2022)Wortsman, Ilharco, Kim, Li, Kornblith, Roelofs, Lopes, Hajishirzi, Farhadi, Namkoong, et~al.]{wortsman2022robust}
Mitchell Wortsman, Gabriel Ilharco, Jong~Wook Kim, Mike Li, Simon Kornblith, Rebecca Roelofs, Raphael~Gontijo Lopes, Hannaneh Hajishirzi, Ali Farhadi, Hongseok Namkoong, et~al.
\newblock Robust fine-tuning of zero-shot models.
\newblock In \emph{CVPR}, 2022.

\bibitem[Xiao et~al.(2010)Xiao, Hays, Ehinger, Oliva, and Torralba]{xiao2010sun}
Jianxiong Xiao, James Hays, Krista~A Ehinger, Aude Oliva, and Antonio Torralba.
\newblock Sun database: Large-scale scene recognition from abbey to zoo.
\newblock In \emph{CVPR}, pages 3485--3492. IEEE, 2010.

\bibitem[Yao et~al.(2021)Yao, Zhang, Zhang, Liu, Chua, and Sun]{yao2021cpt}
Yuan Yao, Ao~Zhang, Zhengyan Zhang, Zhiyuan Liu, Tat-Seng Chua, and Maosong Sun.
\newblock Cpt: Colorful prompt tuning for pre-trained vision-language models.
\newblock \emph{arXiv:2109.11797}, 2021.

\bibitem[Zaken et~al.(2021)Zaken, Ravfogel, and Goldberg]{zaken2021bitfit}
Elad~Ben Zaken, Shauli Ravfogel, and Yoav Goldberg.
\newblock Bitfit: Simple parameter-efficient fine-tuning for transformer-based masked language-models.
\newblock \emph{ACL}, 2021.

\bibitem[Zhang et~al.(2020)Zhang, Sax, Zamir, Guibas, and Malik]{zhang2020side}
Jeffrey~O Zhang, Alexander Sax, Amir Zamir, Leonidas Guibas, and Jitendra Malik.
\newblock Side-tuning: a baseline for network adaptation via additive side networks.
\newblock In \emph{ECCV}, pages 698--714. Springer, 2020.

\bibitem[Zhang et~al.(2021)Zhang, Fang, Gao, Zhang, Li, Dai, Qiao, and Li]{zhang2021tip}
Renrui Zhang, Rongyao Fang, Peng Gao, Wei Zhang, Kunchang Li, Jifeng Dai, Yu~Qiao, and Hongsheng Li.
\newblock Tip-adapter: Training-free clip-adapter for better vision-language modeling.
\newblock \emph{arXiv:2111.03930}, 2021.

\bibitem[Zhou et~al.(2022{\natexlab{a}})Zhou, Yang, Loy, and Liu]{Zhou2021}
Kaiyang Zhou, Jingkang Yang, Chen~Change Loy, and Ziwei Liu.
\newblock Learning to prompt for vision-language models.
\newblock \emph{IJCV}, 2022{\natexlab{a}}.

\bibitem[Zhou et~al.(2022{\natexlab{b}})Zhou, Yang, Loy, and Liu]{Zhou2022}
Kaiyang Zhou, Jingkang Yang, Chen~Change Loy, and Ziwei Liu.
\newblock Conditional prompt learning for vision-language models.
\newblock \emph{CVPR}, 2022{\natexlab{b}}.

\end{thebibliography}
\newpage
\renewcommand{\thesubsection}{\thesection.\alph{section}}
\section{Details of the datasets}
\label{appendix:data}

\Cref{data_stats} gives an overview of the downstream datasets used for the evaluation of our method, including the text prompt templates used to generate classifiers for CLIP.

\begin{table}[!htb]
\setlength\tabcolsep{0.1em}
\begin{center}
\begin{small}
\begin{tabular}{lccccc}
\toprule
Dataset & \#\,Train & \#\,Val. & \#\,Test& Classes & Text Prompt \\
\midrule
CIFAR100 & 50K & - & 10K & 100 & ``This is a photo of a \{ \}''\\
CIFAR10 & 50K & - & 10K & 10 & ``This is a photo of a \{ \}''\\
Flowers102 & 4K & 1.6K & 2.5K & 102 & ``This is a photo of a \{ \}''\\
Food101 & 50K & 20K & 30.3K & 101 & ``This is a photo of a \{ \}''\\
EuroSAT & 13.5K & 5.4K & 8.1K & 10 & ``This is a photo of a \{ \}''\\
SUN397 & 15.9K & 4K & 19.9K & 397 & ``This is a photo of a \{ \}''\\
UCF101 & 7.6K & 1.9K & 3.7K & 101 & ``This is a photo of a \{ \}''\\
SVHN & 73.3K & - & 26K & 10 & ``This is a photo of a \{ \}''\\
OxfordPets & 2.9K & 736 & 3.6K & 37 & ``This is a photo of a \{ \}''\\
DTD & 2.8K & 1.1K & 1.6K & 47 & ``This is a photo of a \{ \}''\\
Resisc45 & 18.9K & 6.3K & 6.3K & 45 & ``This is a photo of a \{ \}''\\
\bottomrule
\end{tabular}
\end{small}
\end{center}
\vspace{0.4em}
\caption{Description of the datasets and the corresponding text prompt used for CLIP. The data is adapted from Bahng \etal \cite{Bahng2022}.}
\label{data_stats}
\vskip -0.1in
\end{table}

\begin{table}[h!]
\setlength\tabcolsep{0.2em}
\begin{center}
\small
\begin{tabular}{cccccc}
\toprule
Table/Figure & Epochs & Number of prompts & TL Size & PGN resolution \\
\midrule
Table 1 & 1000 & 16 & 256 & $224\times224$ \\
Table 2 & 500 & 8 & 64 & $64\times64$\\
% Table 3 & 500 & - & - & -\\
Table 3* & 1000 & 16 & 256 & $224\times224$\\
Table 4 & 500 & 16 & 256 & $224\times224$\\
Figure 2 & 1000 & 16 & 256 & $224\times224$\\
Figure 3 & 500 & 8 & 64 & $64\times64$\\
Figure 4 & 1000 & 16 & 256 & $224\times224$\\
Figure 5 & 1000 & 16 & 256 & $224\times224$\\
\bottomrule
\end{tabular}
\end{center}
\vspace{0.4em}
\caption{Experimental settings for each of our tables and figures. *In Table 3 of the main paper, the reported numbers for VP do not come from our own experimentation, hence our settings do not apply. 
}
\label{tab:exp-details}
% \vskip -0.1in
\end{table}

\section{Additional experimental settings}
\label{appendix:data}

\paragraph{Training Details.}
In~\Cref{tab:exp-details}, we show the training details of the experiments in the main paper. We train the PGN with a learning rate of $0.1$
and apply a cosine decay learning schedule ending at zero learning rate
with a linear warmup for the first $50$ epochs.
We use an SGD optimizer with $0.9$ momentum.
Except when specified,  we use a batch size of $128$ images on one Nvidia-1080TI GPU.
Compared to the $1{,}000$ epochs of concurrent work~\citep{Bahng2022}, we train our network for $500$ epochs by default in the motivation and ablation sections and for $1{,}000$ in the large-scale comparisons (Table 3 of the main paper).

\paragraph{Architectures.}
In~\Cref{table:arch_resnet10},
we show the details of ResNet10 architectures.

\begin{table}[h!]
\begin{center}
	\begin{tabular}{c|c|c}
		\toprule
		stage   &  specification  & $ \begin{matrix} \text{output sizes}\\ H\times W\times C \end{matrix}$ \\ \hline 
		input data & - & $ 224^2\times 3$ \\ \hline 
		$\text{conv}_1$  & $\begin{matrix} 7\times 7, {\color{red} 16}\\ \text{stride } 2, 2\end{matrix}$ & $112^2\times{\color{red} 16}$ \\ \hline 
		$\text{pool}_1$    & $\begin{matrix} 3\times 3, {\color{red} 16}\\ \text{stride } 2, 2\end{matrix}$ & $56^2\times{\color{red} 16}$ \\ \hline 
		$\text{res}_2$     & $\begin{bmatrix} 3\times3, {\color{red} 16}\\3\times3, {\color{red} 16} \end{bmatrix}\times {\color{red} 1}$ & $56^2\times{\color{red} 16}$ \\ \hline 
		$\text{res}_3$     & $\begin{bmatrix} 3\times3, {\color{red} 32}\\3\times3, {\color{red} 32} \end{bmatrix}\times {\color{red} 1}$ & $28^2\times{\color{red} 32}$ \\ \hline 
		$\text{res}_4$     & $\begin{bmatrix} 3\times3, {\color{red} 64}\\3\times3, {\color{red} 64} \end{bmatrix}\times {\color{red} 1}$ & $14^2\times{\color{red} 64}$ \\ \hline 
		$\text{res}_5$     & $\begin{bmatrix} 3\times3, {\color{red} 128}\\3\times3, {\color{red} 128} \end{bmatrix}\times {\color{red} 1}$ & $7^2\times{\color{red} 128}$ \\ \hline 
		$\text{pool}_2$    & $\begin{matrix} 7\times 7, {\color{red} 128}\\ \text{stride } 1, 1\end{matrix}$ & $1^2\times{\color{red} 128}$ \\ 
		\bottomrule
	\end{tabular}
\end{center}
\vspace{0.4em}
	\caption{The structure of ResNet10,
which is modified from ResNet18 to be more light-weight. 
Modifications are marked in {\color{red} red}. 
Note that the final classification layer is omitted.
}\label{table:arch_resnet10}
\end{table}

\paragraph{Feature similarities computation.}
For~Figure~3 in the main paper, we embed the validation set of CIFAR-100 using the three visual encoders of PGN (only), CLIP, and PGN+CLIP. 
For this we cluster the features into 100 clusters using k-means.
After this, the representations can be easily compared with each other using the normalised mutual information score.

\section{Qualitative analysis} %%%%%%%%%%%%%%%%%%%%%%%%%%%%%%%%%%%%%%%%%%%%%%%%%%%%%%
From Table 1 in the paper, we observed that the CLIP zero-shot and the PGN backbone model's performance on their own are low with 63-64\%.
However, when combined, we reach performance increases of +15\% yielding up to 79\% on CIFAR100. 
In this section, we analyse how the simple mechanism behind PGN is allowing the combined model to achieve superior performances.

\begin{figure}[!htb]
\begin{center}
\includegraphics[width=0.68\textwidth]{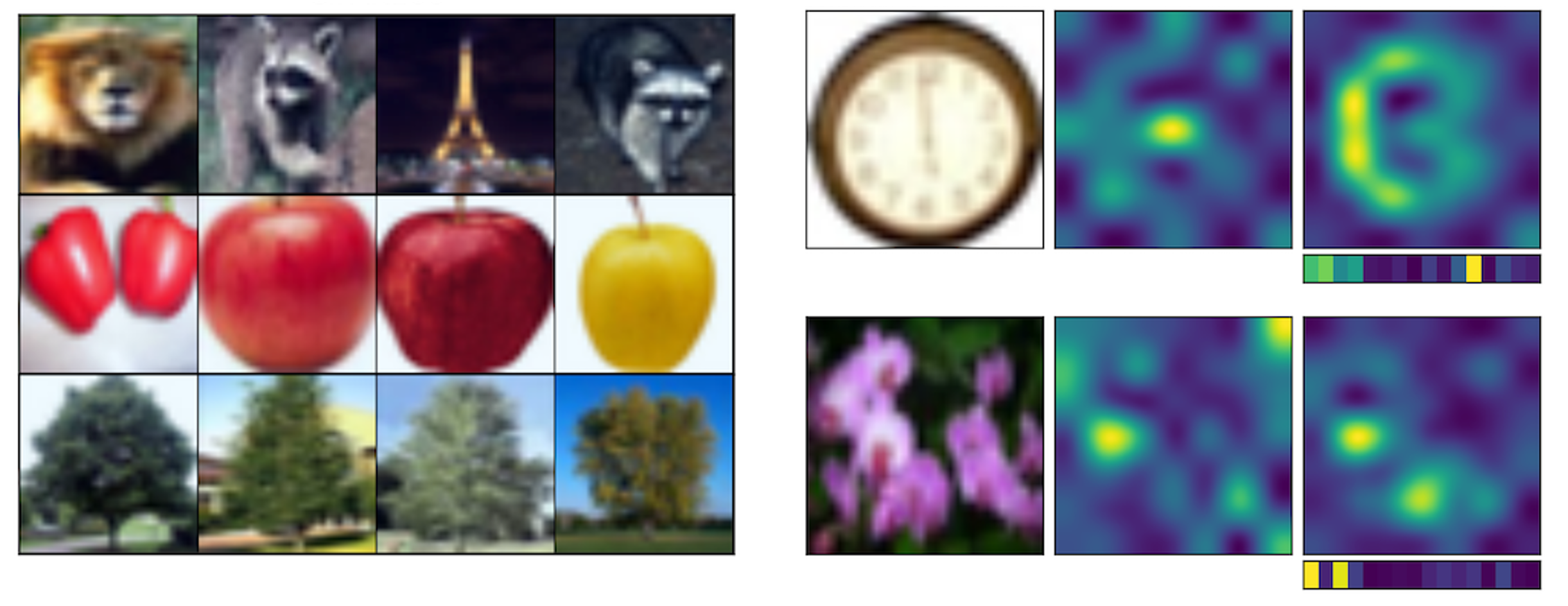}
\end{center}
\caption{\textbf{Token Library items and attention values.} On the left, we show 4 CIFAR100 samples that maximally activate one of three selected items in the token library. Each row in the grid corresponds to one token library item. On the right, we show the individual attention values from the \texttt{CLS} token to the supplied prompts of PGN and IIP. We find that while PGN has an overall lower average attention, the input-dependency successfully yields a wider distribution in adapting the original model.
}
\label{fig:qualitative}
\end{figure}

\paragraph{What do the individual Token Library items stand for?}
To answer this question, we pass the validation sets through the trained PGN model and pick individual tokens that we wish to visualize.
We then pick the top four input samples that have the highest softmax values for the selected item.
The result is shown in~\Cref{fig:qualitative} for CIFAR100.
We find that while some tokens are fairly category specific, such as those for a tree or an apple, some cover much broader categories such as lighting conditions or even geometric structures.
Note however that the PGN is not doing the heavy-lifting in terms of classifying the images by itself, as its output is not well-aligned with the ground-truth, as demonstrated in Figure 3 of the main paper. 
It rather supplies the frozen transformer model with orthogonal information that helps the task. More examples are provided at the end of this document.
\paragraph{How is the computation changed by PGN prompts?}
Next, we analyse the effect of the PGN prompts to the internal computations of the frozen vision transformer.
In~\Cref{fig:qualitative}, we visualize the \texttt{CLS} token's attention map at the final layer of the transformer with or without our PGN.
Despite showing the effect of the prompts on the \textit{last} layer's attention map, we still find a strong effect of the PGN's additionally supplied prompts.
While the effect is not interpretable for low-resolution datasets such as CIFAR, for Pets and Resisc we observe an increased focus on the foreground.
We also show the attention values of the \texttt{CLS} to the 16 supplied prompts below the PGN-CLIP attention maps.
A strong variance between images is seen, demonstrating that the method learns and leverages the input-dependency of the prompts that are supplied to the frozen model. More examples are provided at the end of this document.

\section{Multi-dataset PGN}
We retain the same setting as in our large-scale experiments and train with batches that contain samples from the four datasets in~\Cref{tab:multi}.
The model is thus forced to allocate the token library items in a manner that best supports this task, reducing the overall number of additionally adapted parameters by 75\%.
From~\Cref{tab:multi}, we find that despite this reduction in parameters, the overall performance only decreases by a small amount of 3.7\%, despite the fact that the classification problem is now 193-way and thus much more difficult.
\begin{table}[h]
% \begin{table}[h]
\setlength\tabcolsep{0.55em}
\begin{center}
\begin{small}
\begin{tabular}{r|cccc|ccc}
\toprule
 Method & \scriptsize EurSAT & \scriptsize UCF  & \scriptsize Pets & \scriptsize RESISC & Avg.  & $\Delta$ & $\Sigma$\,params \\
\midrule
CLIP+TP (I) & 40.0 & 59.9 & 85.9 & 42.4 & 57.1 & \multirow{2}{*}{-7.7\%}& - \\ % (40+59.9 + 85.9 +42.4)/4
CLIP+TP {\color{blue}(J)} & 4.4 & 59.7 & 85.8 & 47.7 & 49.4  & & -\\ % (4.4+59.7+85.8+47.7)/4=49.4
\midrule
 \,\,+ PGN (I) & 98.0 & 77.6 & 91.5 & 92.1 & 89.8  & \multirow{2}{*}{-3.7\%}& 5M \\ % (98+77.6+91.5+92.1)/4
 \,\,+ PGN {\color{blue}(J)} & 96.9 & 72.7 & 89.0 & 85.7 & 86.1 & & 1M \\ % (96.9+72.7+89+85.7)/4 = 86.1
\bottomrule
\end{tabular}
\end{small}
\end{center}
\vspace{0.4em}
\caption{\textbf{Training multi-dataset PGN.} 
Adapting and inferring jointly {\color{blue}(J)} over multiple datasets compared to individual (I), per-dataset training and evaluation.
%training one jointly ({\color{blue}J}) PGN for each dataset with training a single PGN individually (I) across four datasets. 
Giving more text prompts (TP) for joint inference leads to a strong decrease in accuracy, yet, joint training of the PGN retains a strong performance while reducing the number parameters by  75\%.
\label{tab:multi}} 

% \end{table}
\vskip -0.3in
\end{table}

\section{Details of the feature similarity analysis}
In the NMI analysis in Figure~3 and~Sec.~4.1 of the main paper, we measure the pairwise alignment between the outputs of the visual encoders we use and the ground truth. 
These are: 
the frozen CLIP model's visual encoder that outputs \texttt{CLS} embedding, 
the trained PGN model that outputs prompts (the $\hat{\mathbf{h}}_V$ in Eqn.~4),
and the combined CLIP+PGN model which uses PGN prompts to modify CLIP's visual encoder's outputs (that outputs \texttt{CLS} embedding after CLIP).
For this, we apply $k$-means clustering to the set of embeddings generated by each encoder individually, setting $k$ equal to the number of ground-truth classes. 
For our experiment, we use the full CIFAR100 test split. 
This yields a set of $3$ pseudo labelings of the dataset. 
After combination with the ground-truth labels, we can make $6$ pairwise comparisons and calculate the normalised mutual information, which measures a generalized correlation between labelings.
The results are shown in~\Cref{appendix_nmi}.

% \HTD{To Jochem: you can use this table if you want, add captions etc.}

\begin{table}[h!]
\begin{center}
\small
\begin{tabular}{c|cccc}
\toprule
\textit{NMI}      & {GT}  & {PGN}    & {CLIP}   & {PGN+CLIP} \\ \hline
{GT}       & 100 & 29.5   & 58.1   & \textbf{70.1}        \\ %\hline
{PGN}      &     & 100    & 27.5   & 33.8        \\ %\hline
{CLIP}     &     &        & 100    & 61.2        \\ %\hline
{PGN+CLIP} &     &        &        & 100     \\ %\hline
\bottomrule
\end{tabular}
\end{center}
\vspace{0.4em}
\caption{Normalized Mutual Information (NMI) score in \%.}
\label{appendix_nmi}
\end{table}

\section{Large-scale comparisons}
In Table 3 in the main paper, the results for linear finetuning are adopted from the original CLIP paper \cite{Radford2021}, whereas the results for full finetuning are taken from VP \cite{Bahng2022}.
\section{Additional experiments}

\paragraph{Comparison between linear and non-linear layer.}
In~\Cref{appendix_abl_projection} we evaluate replacing the final linear layer of $g_\theta$ with a MLP with 1 hidden layer, which allows for a nonlinear mapping between image features and the logits that give rise to the combination coefficients in~Eqn.~3. No significant performance gain is observed.
\begin{table}[h!]
    \begin{center}        
    \setlength\tabcolsep{0.2em}
    \small
    \begin{tabular}{lcc}
        \toprule
        Type & CIFAR100 & SUN397 \\
        \midrule
        % Direct & \\
        % TL \\
        Linear & 77.9 & \textbf{70.5} \\
        MLP & \textbf{78.2} & 70.4 \\
        \bottomrule
    \end{tabular}
    % \label{abl_projection}
    \end{center}
    \vspace{0.4em}
    \caption{Feature projection layer type.}
    \label{appendix_abl_projection}
\end{table}

\paragraph{Unfreezing the classification layer.}
So far, we have utilized CLIP's text prompts (TP) to generate the fixed weights of a linear classifier. In \Cref{abl_classifier}, we compare this approach to a trainable classifier, which takes the TP weights as a starting point.
\begin{table}[h!]
    \begin{center}
    \setlength\tabcolsep{0.2em}
    \begin{small}
    \begin{tabular}{lcc}
        \toprule
        Cls. & CIFAR100 & SUN397 \\
        \midrule
        TP & 79.3 & 70.9 \\
        \,\,+ SGD & 79.3 & 70.3 \\
        \bottomrule
    \end{tabular}
    \end{small}
    \end{center}
    \vspace{0.4em}
    \caption{Training a linear layer in addition to PGN.}
    \label{abl_classifier}
\end{table}
\begin{table}[h!]
    \begin{center}
    \setlength\tabcolsep{0.5em}
    \small
    \begin{tabular}{l|c|cccc}
        \toprule
        Method & ImageNet & A & R & V2 & Sketch \\
        \midrule
        PGN & 66.0 & 22.8 & 62.5 & 56.7 & 36.5 \\
        LP & 67.0 & 10.6 & 38.1 & 1.0 & 36.1 \\
        \bottomrule
    \end{tabular}
    % \label{abl_projection}
    % \vskip -0.1in
    \end{center}
    \vspace{0.4em}
    \caption{Evaluation accuracies on 4 robustness benchmarks. We compare adaptation with a PGN to linear finetuning (LP). We observe that PGN retains much higher scores on these robustness evaluations.}
    \label{robustness}
\end{table}
\paragraph{Experiments on robustness.}
We evaluate the robustness of PGNs by training for 100 epochs onf ImageNet \cite{Deng2009} and evaluating on four ImageNet variations (ImageNet-A \cite{hendrycks2019nae}, ImageNet-R \cite{hendrycks2021many}, Imagenet-V2 \cite{recht2019imagenet} and Imagenet-Sketch \cite{wang2019learning}). For these experiments, we use identical PGN settings as in Table 3 in the paper. The results are shown in Table \ref{robustness} and compared to the case of linear finetuning on the same, frozen CLIP backbone (ViT-B/32).
We see that the PGN outperforms linear finetuning on all robustness benchmark, despite being comparable in terms of its performance on the upstream dataset.
\begin{figure}[t!]
\begin{center}
\includegraphics[width=0.93\textwidth]{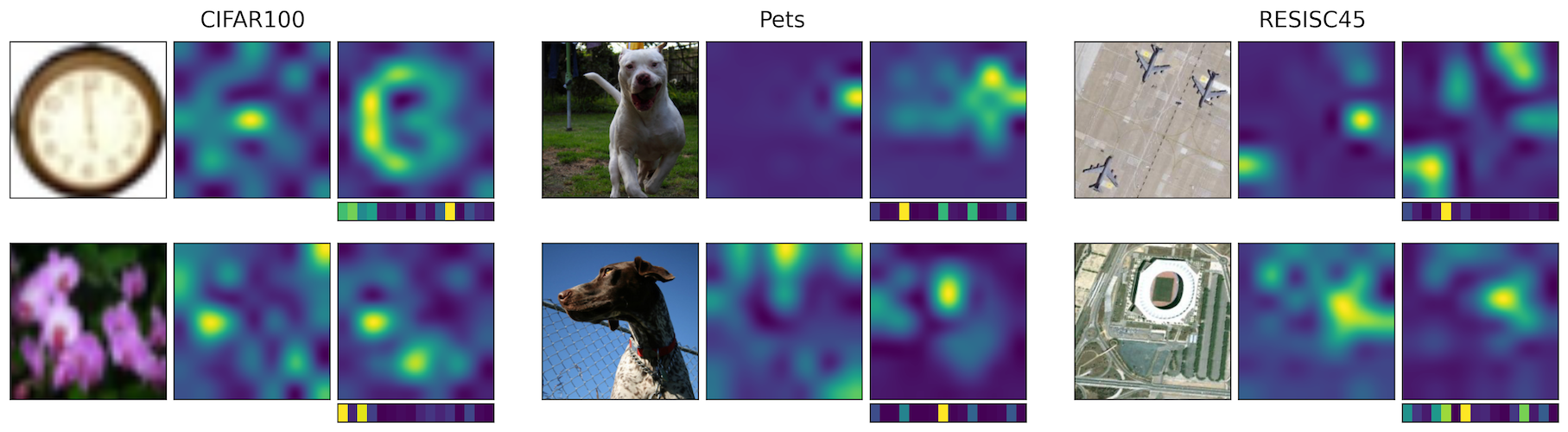}
\end{center}
\caption{{\small
\textbf{Modification of \texttt{CLS} attention maps.} 
We show the attention map of the \texttt{CLS} token for various inputs (left) with the spatial patches for both the original CLIP model's (middle) and the PGN-modified CLIP's final layer (right). 
Below the PGN attention map, we show the attention to PGN's additional prompts.
% We show two samples each from CIFAR100, Pets and Resisc.
We observe a clear modification of the attention map as well as the diverse activation patterns to the supplied tokens.
}}
\label{fig:attention}
\end{figure}
\begin{figure}[t!]
\begin{center}
\includegraphics[width=0.93\textwidth]{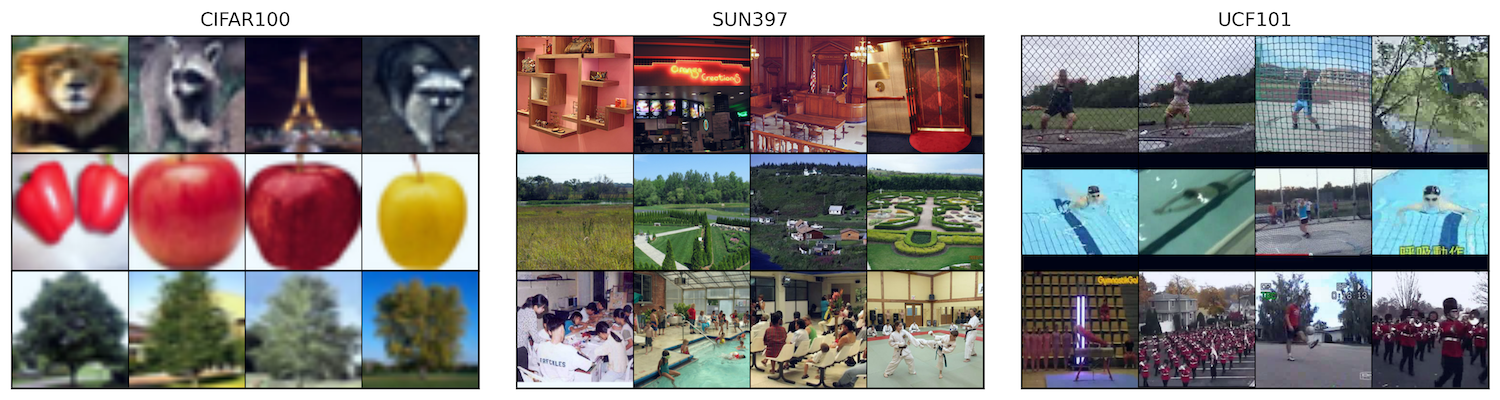}
\end{center}
\caption{{\small\textbf{Token Library example items.} We show 4 samples that maximally activate one of three selected items in the token library for three datasets.
Each row in the grid corresponds to one token library item.
We find that the items can stand for whole objects such as apples and trees for CIFAR100, and also for lower level features such as light warmth or net structures as in UCF101.}}
\label{fig:library}
\end{figure}
\end{document}